\documentclass[10pt,twocolumn,letterpaper]{article}

\usepackage{iccv}
\usepackage{times}
\usepackage{epsfig}
\usepackage{graphicx}
\usepackage{amsmath}
\usepackage{amssymb}
\usepackage{mathtools}
\usepackage{amsthm}
\usepackage{bbding}
\usepackage{xcolor}
\usepackage{multirow}
\usepackage{wrapfig}
\usepackage{microtype}
\usepackage{graphicx}
\usepackage{subfigure}
\usepackage{booktabs} 
\usepackage{adjustbox}
\usepackage{caption}
\usepackage{float}
\usepackage{color}
\usepackage[accsupp]{axessibility}

\usepackage[pagebackref=true,breaklinks=true,letterpaper=true,colorlinks,bookmarks=false]{hyperref}

\iccvfinalcopy 

\def\iccvPaperID{9201} 

\ificcvfinal\fi

\begin{document}

\title{Name Your Colour For the Task: Artificially Discover Colour Naming via  Colour Quantisation Transformer}


\author{Shenghan Su\textsuperscript{1},
    Lin Gu\textsuperscript{3,2}\footnotemark[1]\:,
    Yue Yang\textsuperscript{1,4},
    Zenghui Zhang\textsuperscript{1},
    Tatsuya Harada\textsuperscript{2,3}\\
    \textsuperscript{1}Shanghai Jiao Tong University, \textsuperscript{2}The University of Tokyo,  
    \textsuperscript {3}RIKEN AIP,
    \textsuperscript {4}Shanghai AI Laboratory\\
    \footnotesize{\texttt{\{su2564468850, yang-yue, zenghui.zhang\}@sjtu.edu.cn, lin.gu@riken.jp, harada@mi.t.u-tokyo.ac.jp}}
    }


\twocolumn[{%
\maketitle
\vspace{-5mm}
\begin{figure}[H]
\hsize=\textwidth 
\centering
\includegraphics[width=1.0\textwidth]{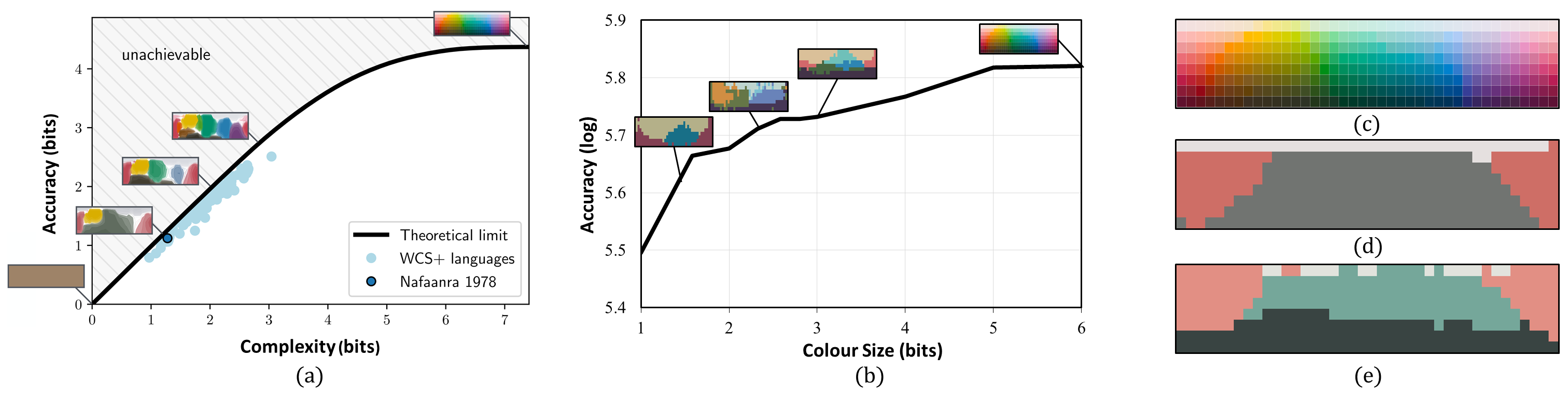}
  \vspace{-7mm}
 \caption{(a) The theoretical limit of efficiency for colour naming (black curve) and  cases of the WCS probability map of human colour language copied from \cite{Zaslavsky2022evolution}. (b) The colour size (from 1-bit to 6-bit)-accuracy curve on the tiny-imagenet-200~\cite{le2015tiny} dataset. The WCS probability maps generated by our CQFormer are also shown along the curve. (c) The colour naming stimulus grid used in the WCS~\cite{kay2009world}. (d) The three-term WCS probability map of CQFormer after embedding 1978 Nafaanra three-colour system ((light (`fiNge'), dark (`wOO'), and warm or red-like (`nyiE')) into the latent representation. (e) The four-term WCS probability map of CQFormer evolved from (d). The evolved fourth colour, yellow-green, is consistent with the prediction of basic colour term theory~\cite{berlin1969basic}.}
 \label{fig:overview}
\end{figure}
}]
\renewcommand{\thefootnote}{\fnsymbol{footnote}} 
\footnotetext[1]{Corresponding author.} 
\ificcvfinal\thispagestyle{empty}\fi



\begin{abstract}
The long-standing theory that a colour-naming system evolves under dual pressure of efficient communication and perceptual mechanism is supported by more and more linguistic studies, including analysing  four decades of diachronic data from the Nafaanra language. This inspires us to explore whether machine learning could evolve and discover a similar colour-naming system via optimising the communication efficiency represented by high-level recognition performance. Here, we propose a novel colour quantisation transformer, CQFormer, that quantises colour space while maintaining the accuracy of machine recognition on the quantised images. Given an RGB image, Annotation Branch maps it into an index map before generating the quantised image with a colour palette; meanwhile the Palette Branch utilises a key-point detection way to find proper colours in the palette among the whole colour space. By interacting with colour annotation,  CQFormer is able to balance both the machine vision accuracy and colour perceptual structure such as distinct and stable colour distribution for discovered colour system. Very interestingly, we even observe the consistent evolution pattern between our artificial colour system and basic colour terms across human languages. Besides, our colour quantisation method also offers an efficient quantisation method that effectively compresses the image storage while maintaining high performance in high-level recognition tasks such as classification and detection. Extensive experiments demonstrate the superior performance of our method with extremely low bit-rate colours, showing potential to integrate into quantisation network to quantities from image to network activation.  The source code is available at \url{https://github.com/ryeocthiv/CQFormer}
\end{abstract}


\section{Introduction}

\leftline{\textit{Hath not a Jew eyes? Hath not a Jew hands, organs,}}
\leftline{\textit{dimensions, senses, affections, passions?}}
\vspace{-3mm}
\noindent \rule[0pt]{\columnwidth}{0.05em}
\rightline{William Shakespeare  ``The Merchant of Venice"}

Does artificial intelligence share the same perceptual mechanism for colours as human beings? We aim to explore this intriguing problem from machine learning perspective.
 
Colour involves the visual reception and neural registering of light stimulants when the spectrum of light interacts with cone cells in the eyes. Physical specifications of colour also include the reflective properties of the physical objects, geometry incident illumination, \textit{etc.}  By defining a colour space~\cite{forsyth2002computer}, people could identify colours directly according to these quantifiable coordinates.   
 


Compared to the pure physiological nature of hue categorisation, the complex phenomenon of colour naming or colour categorisation spans multiple disciplines, from cognitive science to anthropology. Solid diachronic research~\cite{berlin1969basic}  also suggests that human languages constantly evolve to acquire new colour names, resulting in an increasingly fine-grained  colour naming system. This evolutionary process is hypothesised to be under the pressure of communication efficiency and perceptual structure. Communication efficiency requires shared colour partitioning to be communicated accurately with a lexicon as simple and economical as possible. Colour perceptual structure is relevant to human  perception in nature. For example, the colour space distance between nearby colours should correspond to their perceptual dissimilarity. This structure of perceptual colour space has long been used to explain colour naming patterns across languages. A recent analysis of human colour naming systems, especially in Nafaanra, contributes very direct evidence to support this hypothesis through the employment of Shannon’s communication model~\cite{shannon1948mathematical}. Interestingly, this echos the research on colour quantisation, which quantises colour space to reduce the number of distinct colours in an image.



Traditional colour quantisation methods~\cite{mediancut,octree,mediancut_with_dither} are \textit{perception-centred} and generate a new image that  is as visually perceptually similar as possible to the original image. These methods group similar colours in the colour space and represent each group with a new colour, thus naturally  preserving  the perceptual structure. Instead of prioritising the perceptual quality,  Hou \textit{et al.}~\cite{hou2020learning}  proposed a \textit{task-centred/machine-centred} colour quantisation method, ColorCNN, focusing on   maintaining  network classification accuracy in the restricted colour spaces. While achieving an impressive classification accuracy on  even a few-bit image, ColorCNN only identifies and preserves  \textit{machine-centred} structure without directly clustering similar colours in the colour space. Therefore, this pure  \textit{machine-centred} strategy sacrifices perceptual structure and often associates nearby colours with different quantised indices.



Zaslavsky \textit{et al.}~\cite{Zaslavsky2022evolution} measure the communication efficiency in colour naming by analysing the informational complexity based on the information bottleneck (IB) principle. Here, we argue that the network recognition accuracy also reflects the communication efficiency when the number of colours is restricted. Since the human colour naming is shaped by both perception structure and communication efficiency~\cite{Zaslavsky19a}, we integrate the need for both \textit{perception} and \textit{machine} to propose a novel end-to-end colour quantisation transformer, CQFormer, to discover the artificial colour naming systems. 


As illustrated in Fig.~\ref{fig:overview} (b), the recognition accuracy increases with the number of colours in our discovered colour naming system. Surprisingly, the complexity-accuracy trade-offs are similar to the numerical results (Fig.~\ref{fig:overview} (a)) independently derived from linguistic research ~\cite{Zaslavsky2022evolution}. What is more, after embedding 1978 Nafaanra three-colour system (Nafaanra-1978)  into the latent representation of CQFormer (Fig.~\ref{fig:overview} (d)), our method  automatically evolves the fourth colour closed to yellow-green, matching the basic colour terms theory~\cite{berlin1969basic} summarised  in  different languages. Berlin and Kay found universal restrictions on colour naming  across cultures and claimed languages acquire new basic colour categories  in a strict chronological sequence. For example, if a culture has three colours (light (`fiNge'), dark (`wOO'), and warm or red-like (`nyiE') in Nafaanra), the fourth colour it evolves should be yellow or green, exactly the one  (Fig.~\ref{fig:overview} (e))  discovered by our CQFormer.

\begin{figure*}[h!]

 \centering
 \includegraphics[width=\linewidth]{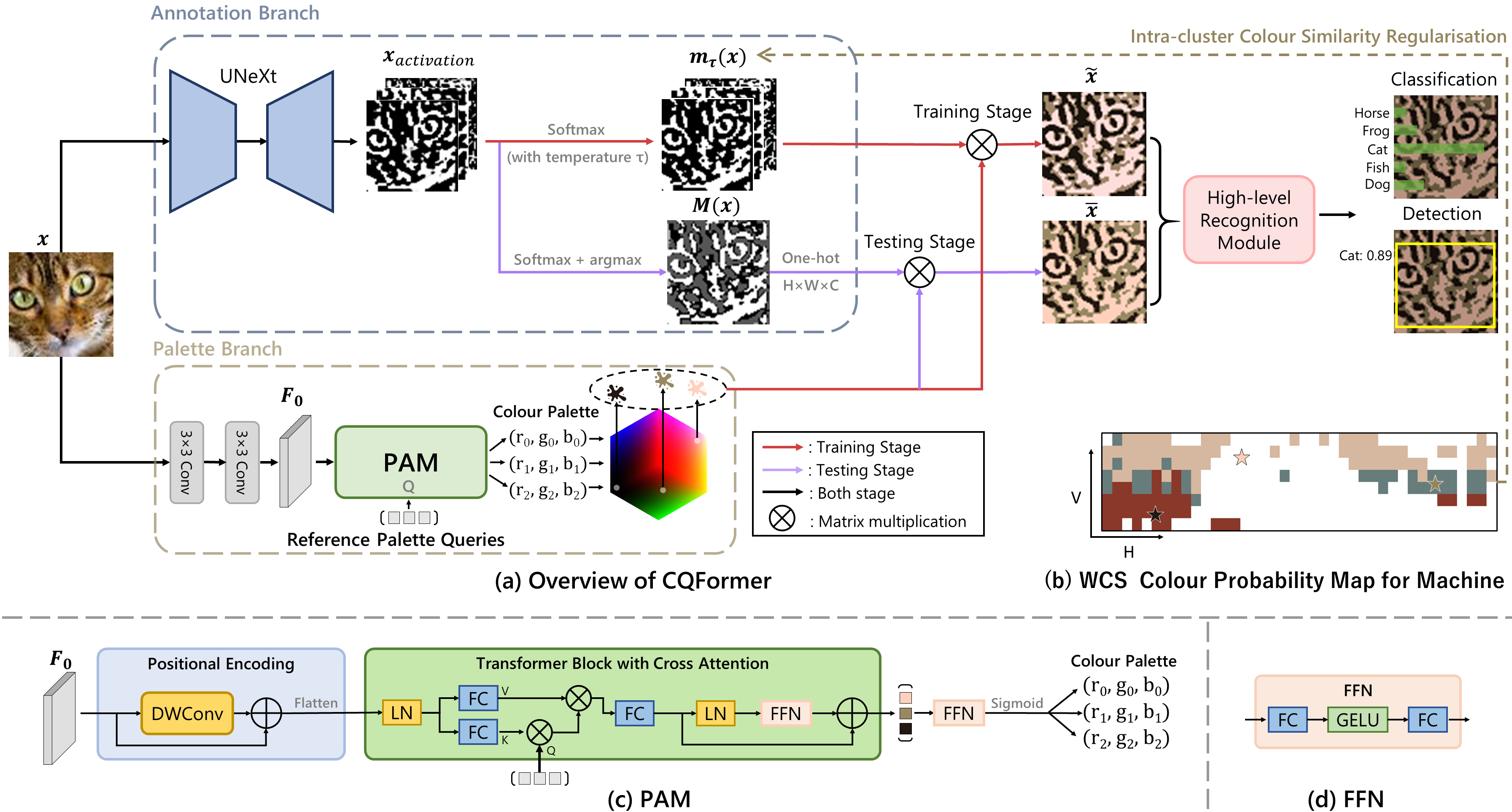}
 \vspace{-0.5cm}
 \caption{(a) Overview of our CQFormer (taking 3-colour quantisation as an example). Red lines are only for training stage, purple lines are only for testing stage, and black lines are both used in both stage. (b) WCS colour probability map counted across all pixels of the input image. (c) Detailed structure of PAM. (d) Detailed structure of FFN.}
 \label{fig:CQFormer}
\vspace{-0.5cm}
\end{figure*}

The pipeline of CQFormer, shown in Fig.~\ref{fig:CQFormer}, comprises two branches:  Annotation Branch and Palette Branch.  Annotation Branch annotates each pixel of the input RGB image with the proper quantised colour index before painting the index map with the corresponding colour in  the  colour palette. We localise the colour palette  in the whole RGB colour space with a novel Palette Branch, which detects the key-point with explicit attention queries of transformer. During the training stage,  as illustrated in the red line and black line of Fig.~\ref{fig:CQFormer} (a),  Palette Branch interacts with an input image and Reference Palette Queries to maintain the perceptual structure by reducing the perceptual structure loss. This \textit{perception-centred} design groups similar colours and ensures  the  colour palette sufficiently represents the colour naming system defined by  the  World Color Survey (WCS) colour naming stimulus grids. As shown in Fig.~\ref{fig:CQFormer} (b), each item in the colour palette (noted by an asteroid) lies in the middle of the corresponding colour distribution in the  WCS colour naming probability map. Finally, the quantised image is passed to a high-level recognition module for \textit{machine accuracy} tasks such as classification and detection.  Through the joint optimisation of CQFormer and consequent high-level module, we can balance both \textit{perception} and \textit{machine}. Besides automatically discovering the colour naming system, our CQFormer also offers an effective solution to extremely compress image storage while maintaining high performance in high-level recognition tasks. For example, the CQFormer achieves 50.6\% top-1 accuracy on the CIFAR100 \cite{cifar10} dataset with only a 1-bit colour space (\textit{i.e.}, two colours).  The extremely low-bit quantisation of our also demonstrates the potential to integrate into quantisation network research~\cite{MohammadECCV16binary,yang2019quantization}, allowing the end-to-end optimisation from image to weight and activation.

Our contributions could be summarised as follows:
\begin{itemize}
\item We propose a novel end-to-end colour quantisation transformer, CQFormer, to artificially discover a colour naming system by counting both perception and machine. The discovered colour naming system shows a similar pattern to human language on colour. 
\item We propose a novel colour quantisation method taking palette generation as an attention-based key-point detection task. With the input of independent attention queries, it automatically generates 3D coordinates as the selected colour in the whole colour space.
\item Our colour quantisation achieves superior classification and object detection performance with extremely low bit-rate colours.
\end{itemize}

\section{Related Works}
\label{sec:Related Works}

\noindent \textbf{Colour Quantisation.} Colour quantisation, also known as optimal palette generation, compresses images by remapping original pixel colours to a limited set of entries in a small palette.

Traditional colour quantisation~\cite{CQ_of_images,deng1999peer,liang2003general,wu1992color} aims to reduce the colour space while preserving visual fidelity. These \textit{perception-centred} methods usually cluster colours to create a new image as visually perceptually similar as possible to the original image. For example, MedianCut~\cite{mediancut}  and OCTree~\cite{octree} are two representative clustering-based algorithms. Dithering~\cite{dithering} eliminates visual artefacts by including a noise pattern. The colour-quantised images can be expressed as indexed colour~\cite{poynton2012digital}, and encoded with PNG ~\cite{boutell1997png}.

There are also much of  efforts \cite{chang2015palette,cho2017palettenet} to generate the optimal colour palette and  recolour the image. For example, Bahng \textit{et al.} \cite{bahng2018coloring} proposed the text-to-palette generation networks to generate an appropriate palette according to the semantics of the text input. Yoo \textit{et al.} \cite{yoo2019coloring} leverage the memory networks to retrieve the feature of colour palette for colourisation with limited data. Li \textit{et al.} \cite{li2022neural} develop a self-supervised approach to recolouring images from design-oriented fields by reproducing colour palettes with luminance distributions different from the input.


Recently, Hou \textit{et al.} \cite{hou2020learning}  propose a pure \textit{machine-centred} CNN-based colour quantisation method, ColorCNN, which effectively maintains the informative structures under an extremely constrained colour space. In addition to colour quantisation, Camposeco  \textit{et al.}~\cite{camposeco2019hybrid} also design a task-centred image compression method for localisation in 3D map data. Since human colour naming reflects both perceptual structure and communicative need, our CQFormer also considers both \textit{perception} and \textit{machine} to discover the colour naming system artificially.

\noindent \textbf{World Color Survey. }The World Color Survey (WCS)~\cite{kay2009world} comprises colour name data from 110 languages of non-industrialised societies~\cite{Zaslavsky2019b}, \textit{w.r.t.} the stimulus grid shown in Fig.~\ref{fig:overview} (c). There are 320 colour chips in colour naming stimulus grids, and each chip is at its maximum saturation for that hue/lightness combination, while columns correspond to equally spaced Munsell hues and rows to equally spaced Munsell values. Participants were asked to name the colour of each chip to record the colour naming system, generating the WCS probability map for human language (\textit{e.g.} the Nafaanra-1978 in Fig. \ref{fig:inject_human_language} (c)).

\noindent \textbf{Colour Categorisation/Naming. }
Van De Weijer \textit{et al.} \cite{van2007learning} gathered datasets from Google and Ebay with the explicit aim of learning colour names from real-world images. Parraga and Akbarinia\cite{NICE_PLOS} then recruited 17 paid subjects and employed a psychophysical experiment to bridge the gap between the physiology of the visual system and colour categorisation. Gibson \textit{et al.} \cite{gibson2017color} indicated  that warm colours are communicated more efficiently than cool colours in general, and crucially, categorical variations between languages are attributed to differences in the practical value of colour.  Siuda-Krzywicka \textit{et al.}\cite{siuda2019color}  indicates the independence of colour categorisation and naming, while also offering a plausible neural foundation for the process of colour naming. Chaabouni \textit{et al.}\cite{Communicating_pnas}  treated communicative concepts as the primary driving force behind the formation of colour categories and  demonstrated  that continuous message passing increased system complexity and reduced efficiency. De Vries \textit{et al.} \cite{Emergent_eLife} demonstrated that an artificial visual system develops colour categorical boundaries resembling those of humans by learning to recognise objects in images.

\section{Methodology}
\subsection{Problem Formulation}
\label{sec:Problem Formulation}

For a dataset $\mathcal{D}$ containing image-label pairs $(\boldsymbol{x},y)$, the recognition network $f_{\theta}(\cdot)$ takes the input image $\boldsymbol{x}$ and predicts its label $\hat{y}= f_{\theta}(x)$ (\textit{e.g.}, the class probability in a classification task). The network's parameters $\theta$ can be optimised by minimising the loss between the predicted label $\hat{y}$ and the ground truth label $y$, defined as the machine-centred loss $\mathcal{L}_{M}$. This process allows us to find the optimal parameter set $\theta^{\star}$.
\vspace{-0.1cm}
\begin{equation}
\setlength{\abovedisplayskip}{3pt}
\setlength{\belowdisplayskip}{3pt}
\theta^{\star} = \underset{\theta}{\arg \min } \sum_{(\boldsymbol{x}, y) \in \mathcal{D}} \mathcal{L}_{M}\left(y, f_{\theta}(\boldsymbol{x})\right). \label{eq:base}
\end{equation}
\vspace{-0.3cm}


We aim to discover the artificial colour naming system that balances the need for both machine accuracy and human perception. To achieve this, CQFormer focuses not only on recognition accuracy but also on preserving the perceptual structure of the image. The function in Eq.~\ref{eq:base} can be rewritten as follows:
\vspace{-0.1cm}
\begin{equation}
\setlength{\abovedisplayskip}{3pt}
\setlength{\belowdisplayskip}{3pt}
\psi^{\star}, \theta^{\star} = \underset{\psi,\theta}{\arg \min } \sum_{(\boldsymbol{x}, y) \in \mathcal{D}} \mathcal{L}_{M}\left(y, f_{\theta}\left(g_{\psi}(\boldsymbol{x})\right)\right)+ \mathcal{L}_{P}  ,
\end{equation}


\noindent where $\psi$ and $\theta$ represent the parameters of the CQFormer $g$ and the high-level recognition network $f$, respectively. We jointly optimise $g$ and $f$ to find the optimal parameters $\psi^{\star}$ and $\theta^{\star}$. Additionally, $\mathcal{L}_{P}$ is a perceptual structure loss that is focused on preserving the perceptual structure of the image and will be further explained in detail in Sec.~\ref{sec:Perceptual Structure Loss}.

\subsection{CQFormer Architecture}
\label{sec:CQFormer Architecture}


\noindent \textbf{Overall Architecture.} An overview of the CQFormer is illustrated  in Fig.~\ref{fig:CQFormer} (a). It consists of two main branches: (1) Annotation Branch, which assigns a quantised colour index to each pixel of the input RGB image, and (2) Palette Branch, which is  responsible for generating a suitable colour palette.


Annotation Branch of CQFormer takes an original image $\boldsymbol{x} \in \mathbb{R}^{H \times W \times 3}$ as input, where $H$ and $W$ are the height and width of the image, respectively. During the training stage, it generates a probability map $m_{\tau}(\boldsymbol{x}) \in \mathbb{R}^{H \times W \times C}$, where $C$ is the number of quantised colours, and $\tau >0$ is the temperature parameter of the Softmax function~\cite{softmax_temperature}. During the testing stage, it produces a one-hot colour index map $\operatorname{One-hot}(M(\boldsymbol{x})) \in \mathbb{R}^{H \times W \times C}$, where each pixel of the image is assigned a single colour index among the $C$ quantised colours.



Palette Branch of CQFormer takes the original image $\boldsymbol{x}$ and Reference Palette Queries $\textbf{Q} \in \mathbb{R}^{C \times d}$ as input. These queries are composed of $C$ learnable vectors of dimension $d$, each representing an automatically mined colour. The queries $\textbf{Q}$ interact with the keys $\textbf{K} \in \mathbb{R}^{(\frac{HW}{16}) \times d}$ and values $\textbf{V} \in \mathbb{R}^{(\frac{HW}{16}) \times d}$, generated from the input image $\boldsymbol{x}$ to produce the colour palette $P(\boldsymbol{x}) \in \mathbb{R}^{C \times 3}$. This palette consists of $C$ triples $(R,G,B)$, each representing one of the machine-discovered $C$ colours.


Finally, CQFormer produces the quantised image by performing a matrix multiplication between $m_{\tau}(\boldsymbol{x})$ and $P(\boldsymbol{x})$ during the training stage. During the testing stage, it is obtained from $\operatorname{One-hot}(M(\boldsymbol{x}))$ and $P(\boldsymbol{x})$. The quantised image is then fed into the high-level recognition module for object detection and classification tasks.

\noindent \textbf{Annotation Branch.} The first component of Annotation Branch is a UNeXt encoder~\cite{valanarasu2022unext} that generates per-pixel categories. Given the input image $\boldsymbol{x}$, the encoder produces a class activation map $\boldsymbol{x}_{activation} \in \mathbb{R}^{H \times W \times C}$, which contains crucial and semantically rich features.  Then, different approaches are used to process the class activation map during testing and training.

\noindent \textit{(1) Testing stage.} As shown by the purple lines in Fig.\ref{fig:CQFormer} (a), we use the $\boldsymbol{x}_{activation}$ as the input to a Softmax function over $C$ channels, and then apply an $\arg \max$ function to produce a colour index map $M(\boldsymbol{x}) \in \mathbb{R}^{H \times W}$:
\vspace{-0.15cm}
\begin{eqnarray}
M(\boldsymbol{x}) = \underset{C}{\arg \max}(\operatorname{Softmax}(\boldsymbol{x}_{activation})).
\end{eqnarray}
\vspace{-0.4cm}


Subsequently, the colour index map is transformed into a one-hot encoding, denoted as $\operatorname{One-hot}(M(\boldsymbol{x})) \in \mathbb{R}^{H \times W \times C}$, which is then combined with the colour palette $P(\boldsymbol{x})$ through matrix multiplication, resulting in the generation of the test-time colour-quantised image $\Bar{\boldsymbol{x}}$.
\vspace{-0.2cm}
\begin{eqnarray}
\setlength{\abovedisplayskip}{0pt}
\setlength{\belowdisplayskip}{0pt}
\Bar{\boldsymbol{x}} = \operatorname{One-hot}(M(\boldsymbol{x})) \otimes P(\boldsymbol{x}),
\end{eqnarray}
\vspace{-0.6cm}

\noindent where $\otimes$ represents matrix multiplication.


\noindent \textit{(2) Training stage.} As depicted  by the red lines in Fig.\ref{fig:CQFormer} (a), since the $\arg \max$ function is not differentiable, we use the Softmax function as a substitute during the training stage. To prevent overfitting,  we incorporate a temperature parameter $\tau$~\cite{softmax_temperature} into the Softmax function,  pushing the probability map distribution closer to a one-hot vector. This results in the probability map $m_{\tau}(\boldsymbol{x})$ with temperature parameter $\tau$, computed as:
\vspace{-0.1cm}
\begin{eqnarray}
m_{\tau}(\boldsymbol{x})= \operatorname{Softmax}(\frac{\boldsymbol{x}_{activation}}{\tau}).
\end{eqnarray}
\vspace{-0.4cm}



As extensively discussed in \cite{softmax_temperature}, the output is similar to a one-hot vector with large diversity when the temperature is low ($0< \tau < 1$) and a uniform distribution with small diversity otherwise ($\tau > 1$). Therefore, we set $0 < \tau < 1$ and the train-time colour-quantised image $\widetilde{\boldsymbol{x}}$ is generated as:
\vspace{-0.6cm}
\begin{eqnarray}
\widetilde{\boldsymbol{x}} = m_{\tau}(\boldsymbol{x}) \otimes P(\boldsymbol{x}).
\end{eqnarray}  
\vspace{-0.6cm}


\noindent \textbf{Palette Branch.}  We locate the representative colours by using an attention-based key-point detection strategy, which is originally designed to use transformer queries to automatically find the key-point location, such as bounding-boxes~\cite{carion2020detr}, human poses~\cite{li2021pose} and  colour matrix with gamma~\cite{Cui_2022_BMVC}. In other words, we reformulate the problem of colour quantisation into a 3D spatial key-point localisation task within the whole RGB colour space.

Given the input image $\boldsymbol{x}$, we first extract a high-dimensional lower resolution feature $F_{0} \in \mathbb{R}^{\frac{H}{4} \times \frac{W}{4} \times d}$ using two stacked convolution layers. The $F_{0}$ is then passed to the Palette Acquisition Module (PAM) to acquire the colour palette $P(\boldsymbol{x})$. As shown in Fig.~\ref{fig:CQFormer} (c), different from DETR~\cite{carion2020detr}, our $\textbf{Q} \in \mathbb{R}^{C \times d}$, called Reference Palette Queries, are explicit learnable embeddings without extra multi-head self-attention, which attends \textbf{K} and \textbf{V} generated from $F_{0}$. The positional encoding mechanism is a depth-wise convolution \cite{chollet2017xception}, which is suitable for different input resolutions. After that, the position-encoded feature is flattened into sequences before being passed into our transformer block. Here, the \textbf{K} and \textbf{V} are generated by two fully-connection (FC) layers separately, and the cross-attention is estimated as:
\vspace{-0.3cm}
\begin{eqnarray}
\operatorname{Attention}(\textbf{Q}, \textbf{K}, \textbf{V})   =  \operatorname{Softmax}(\frac{\textbf{QK}^{\top }}{\sqrt{d}})\textbf{V}.
\end{eqnarray}
\vspace{-0.3cm}

After a feed forward network (FFN) \cite{dosovitskiy2020vit}, including two FC layers and a GELU \cite{hendrycks2016gaussian} activation function, then with a residual connection \cite{he2016deep}, the Reference Palette Queries $\textbf{Q}$ are transformed into output embeddings. We then decode embeddings into 3D coordinates with another FFN and a Sigmoid function,  resulting in the  final localisation of colour palette $P(\boldsymbol{x}) \in \mathbb{R}^{C \times 3}$ within the whole RGB colour space.


\noindent \textbf{High-Level Recognition Module.}  Taking the  quantised image ($\Bar{\boldsymbol{x}}$ or $\widetilde{\boldsymbol{x}}$) as input, the high-level recognition module predicts the results of object detection or classification.


\subsection{WCS Colour Probability Map}
As introduced in WCS of Sec. \ref{sec:Related Works}, the WCS probability map $m_{\operatorname{WCS}} \in \mathbb{R}^{8 \times 40 \times C}$ for human language is collected by the participants' colour perception in WCS. We also define the WCS colour probability map for machine (see in Fig.~\ref{fig:overview} (d) (e)) and generate it from the datasets. First, the colour index of each pixel is mapped to the grid with the closest (hue, value) coordinate as the pixel. Second, we count the frequency of occurrence of each colour index in each grid, and select the pixel's colour index with the highest frequency as the grid's colour index. Therefore,  $C$ clusters  are formed according to the $C$ grid's colour indexes in WCS colour probability map. Finally, each cluster is shown in the colour corresponding to the centre of mass of its colour category.

The WCS colour probability map for machine is then used to measure the similarity of colours within the same cluster in Sec.\ref{sec:Perceptual Structure Loss}. More importantly, it creates a strong correlation between human language and machine vision since they can be represented in the same format. Therefore, we explore the colour evolution using the WCS colour probability map in Sec.\ref{sec:Colour Evolution}.


\subsection{Perceptual Structure Loss}
\label{sec:Perceptual Structure Loss}

To ensure that CQFormer also produces visually pleasing and structurally sound results, an additional perceptual structure loss $\mathcal{L}_{P}$ is added to the training process:
\vspace{-0.2cm}
\begin{eqnarray}
\mathcal{L}_{P} = \alpha R_{\operatorname{Colour}} +\beta R_{\operatorname{Diversity}} + \gamma \mathcal{L}_{\operatorname{Perceptual}}.
\label{eq:loss_p}
\end{eqnarray}
\vspace{-0.6cm}

$\alpha$ , $\beta$ and $\gamma$ are three non-negative parameters, which combine intra-cluster colour similarity regularisation $R_{\operatorname{Colour}}$, diversity regularisation $R_{\operatorname{Diversity}}$ and perceptual similarity loss $\mathcal{L}_{\operatorname{Perceptual}}$, respectively. Therefore, the objective of training the CQFormer is to minimise the total loss $\mathcal{L_{\operatorname{total}}}$:
\vspace{-0.1cm}
\begin{eqnarray}
\mathcal{L_{\operatorname{total}}} = \mathcal{L}_{M} + \mathcal{L}_{P}. 
\label{eq:loss}
\end{eqnarray}
\vspace{-0.5cm}

\noindent \textbf{Intra-cluster Colour Similarity Regularisation.} The CQFormer associates each pixel of the input image with a colour index and forms $C$ clusters covering different parts of the WCS colour probability map. To ensure that the pixels within the same cluster are perceptually similar in colour, we propose the intra-cluster colour similarity regularisation $R_{\operatorname{Colour}}$, which is the mean of the colour variance 
within the $C$ clusters. At first, the centroid colour values of the $C$ clusters $\{\mu_{1},\mu_{2},\dots,\mu_{C}\}$ are calculated.  Then, we compute $C$ squared colour distances $\operatorname{d}^{2}_{\operatorname{HSV}}$ between all pixels and their centroid colour value in each cluster. Here, the $\operatorname{d}^{2}_{\operatorname{HSV}}$ is calculated in Munsell HSV colour space as:
\begin{eqnarray}
\small
\begin{split}
&\operatorname{d}^{2}_{\operatorname{HSV}}(h_1,s_1,v_1, h_2,s_2,v_2 ) \\ 
&= \left(v_{2}-v_{1}\right)^{2}+s_{1}^{2} v_{1}^{2}+s_{2}^{2} v_{2}^{2}-2 s_{1} s_{2} v_{1} v_{2} \cos \left(h_{2}-h_{1}\right).
\label{hsv_dist}
\end{split}
\end{eqnarray}
\vspace{-0.3cm}

Finally, we take the mean value of the $C$ squared distances as the $R_{\operatorname{Colour}}$:
\vspace{-0.1cm}
\begin{eqnarray}
R_{\operatorname{Colour}}=  \frac{1}{C} \times \sum_{c=1}^{C} \frac{1}{N_c} \sum_{i=1}^{N_c} \operatorname{d}^{2}_{\operatorname{HSV}}(\boldsymbol{x}_{c}[i],\mu_{c}),
\end{eqnarray}
\vspace{-0.2cm}

\noindent where $N_c$ is the number of all pixels in $Cluster_{c}$, and $\boldsymbol{x}_{c}[i]$ represents the Munsell HSV value of the $i$-th pixel.

\noindent \textbf{Diversity Regularisation.} To encourage the CQFormer to select at least one pixel of all $C$ colours, we adopt the diversity regularisation term $R_{\operatorname{Diversity}}$ proposed by Hou \textit{et al.} \cite{hou2020learning}. Diversity is a simple yet efficient metric and serves as an implicit entropy of the assignment distribution. The $R_{\operatorname{Diversity}}$ is calculated as:
\vspace{-0.2cm}
\begin{eqnarray}
\small
R_{\operatorname{Diversity}} = \log _{2} C \times\left(1-\frac{1}{C} \times \sum_{c} \max _{(u, v)}[m_{\tau}(\boldsymbol{x})]_{u, v}\right).
\end{eqnarray}
\vspace{-0.2cm}

\noindent \textbf{Perceptual Similarity Loss.} The perceptual similarity loss, denoted by  $\mathcal{L}_{\operatorname{Perceptual}}$, is a mean squared error (MSE) loss between the quantised image $\widetilde{\boldsymbol{x}}$ and the input image $\boldsymbol{x}$. It ensures that each item of the colour palette $P(\boldsymbol{x})$  lies in the centre of corresponding colour distribution in the WCS colour probability map (noted by asteroids in  Fig.~\ref{fig:CQFormer} (b)). The $\mathcal{L}_{\operatorname{Perceptual}}$ is calculated as:
\vspace{-0.3cm}
\begin{eqnarray}
\mathcal{L}_{\operatorname{Perceptual}} = \mathcal{L}_{\operatorname{MSE}}(\widetilde{\boldsymbol{x}}, \boldsymbol{x}).
\end{eqnarray}
\vspace{-0.6cm}

\subsection{Colour Evolution}
\label{sec:Colour Evolution}

With CQFormer, we explore the colour evolution based on the classification task, involving two successive stages with different loss functions. Since the CQFormer initially has no prior knowledge of colour naming systems associated with corresponding human languages, the first embedding stage aims to embed the colour perceptual knowledge of a certain language into the latent representation of the CQFormer. For example, the CQFormer first learns and matches the 1978 Nafaanra three-colour system by forcing the CQFormer to output a similar WCS colour probability map to that for Nafaanra. Here, we design two embedding solutions and loss functions, \textit{i.e.} $\mathcal{L_{\operatorname{Full-Embedding}}}$ and $\mathcal{L_{\operatorname{Central-Embedding}}}$, to  distil either full colour probability map embedding or only representative colours to CQFormer. The second evolution stage then lets CQFormer evolve more colours, \textit{i.e.} splitting the fourth colour from the learned three-colour system under the pressure of both accuracy and perceptual structure.



\noindent \textbf{Embedding Stage.}

\begin{figure}[htp]
 \centering
 \includegraphics[width=\linewidth]{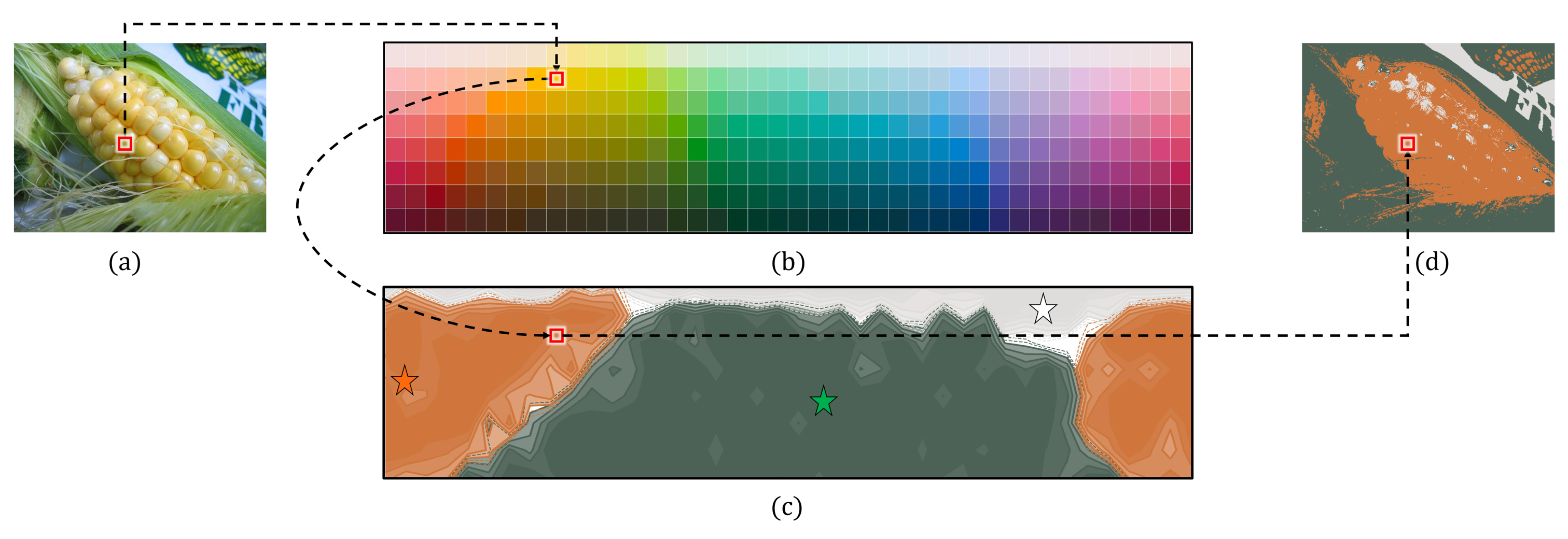}
 \vspace{-0.7cm}
 \caption{The procedure of colour probability map embedding. (a) The origin RGB image. (b) The WCS colour naming stimulus grids. (c) The $m_{\operatorname{WCS}}$ of Nafaanra-1978 copied from \cite{Zaslavsky2022evolution}. (d)  The human language probability  map after $\arg \max$ function for Nafaanra-1978: $\arg \max(m_{\operatorname{Human}}(\boldsymbol{x}))$.}
 \label{fig:inject_human_language}
  \vspace{-0.3cm}
\end{figure}

\noindent \textit{(1) Colour Probability Map Embedding.} As illustrated in Fig.~\ref{fig:inject_human_language}, this embedding forces our CQFormer to match the identical WCS colour probability map of a certain language. At first, we ensure that the CQFormer outputs the same number of quantised colours as the certain human colour system. For each pixel in the input image, we collect its Munsell (hue, value) coordinate and spatial position coordinate $(i,j)$ in the input image. Then we locate this (hue, value) coordinate on the WCS colour probability map of a certain language (Fig.~\ref{fig:inject_human_language} (a)$\rightarrow$(b)$\rightarrow$(c)) to find its corresponding $C$ probability values $p_{\operatorname{Human}}(i,j) \in \mathbb{R}^{C}$ of human colour categories,\textit{e.g.} 9\% for `fiNge', 4\% for `wOO', and 87\% for `nyiE'. After performing the above operations on all pixels of the input image, a set of probability values $\{p_{\operatorname{Human}}(i,j)\}_{(i,j)=(1,1)}^{(H,W)}$ are generated. We arrange each of them according to their spatial position coordinates $(i,j)$ to obtain a new matrix regarded as human language probability map $m_{\operatorname{Human}}(\boldsymbol{x}) \in \mathbb{R}^{H \times W \times C }$. Then we use a cross-entropy loss $\mathcal{L}_{CE}$ instead of $\mathcal{L}_{P}$. Finally, for this full embedding solution, we minimise the loss function as:
\vspace{-0.2cm}
\begin{eqnarray}
\mathcal{L_{\operatorname{Full-Embedding}}} = \mathcal{L}_{M} + \mathcal{L}_{CE}(m_{\tau}(\boldsymbol{x}), m_{\operatorname{Human}}(\boldsymbol{x})). 
\label{eq:loss1}
\end{eqnarray}
\vspace{-0.5cm}


By minimising the above Eq.~\ref{eq:loss1}, our CQFormer inherits the full colour naming system of human language.

\noindent \textit{(2) Central Colour Embedding.}  Alternatively, we could distil less colour naming information. Here, we only embed representative colours and  their $C$ central colour (hue, value) coordinates $\{\mu_{\operatorname{Human},c}\}_{c=1}^{C} $ in the WCS colour probability map of the certain  human language (see three asteroids in Fig.~\ref{fig:inject_human_language} (c)). For this embedding solution, we minimise the loss function as:
\vspace{-0.2cm}
\begin{eqnarray}
\mathcal{L_{\operatorname{Central-Embedding}}} = \mathcal{L}_{M} + R_{\operatorname{Colour}}, 
\label{eq:loss2}
\end{eqnarray}
\vspace{-0.5cm}

\noindent where we replace the $\mu_{c}$ in $R_{\operatorname{Colour}}$ with aforementioned $\mu_{\operatorname{Human},c}$ and ignore the saturation of Munsell HSV by setting $s_1=s_2=1$.

\begin{table*}[htp!]
\renewcommand\arraystretch{1.1}
\caption{Object detection results on MS COCO dataset~\cite{COCO} with Sparse-RCNN~\cite{sun2021sparse} detector, here we report the average precision (AP) value.}
\vspace{-2mm}
\centering
\begin{tabular}{l|ccccccc}
\hline
Method             & 1-bit & 2-bit & 3-bit & 4-bit & 5-bit & 6-bit & Full Colour (24-bit) \\ \hline
Upper bound        & -     & -     & -     & -     & -     & -     & 45.0         \\ \hline
Median Cut (w/o D)~\cite{mediancut} & 11.5   & 12.7   & 15.4  & 17.0  & 20.4  & 23.2  & -             \\ \hline
Median Cut (w D)~\cite{mediancut_with_dither}   & 12.3   & 13.8  & 15.2  & 19.6  & 21.8  & 25.6  & -             \\ \hline
OCTree~\cite{octree}             & 10.7   & 13.4   & 13.2   & 16.7  & 18.9  & 22.8  & -             \\ \hline
\textbf{CQFormer}           & \textbf{13.9}  & \textbf{16.5}  & \textbf{18.8}  & \textbf{21.5}  & \textbf{27.5}  & \textbf{29.8}  & -             \\ \hline
\end{tabular}
\label{Tab:Detection_sparse}
\vspace{-0.2cm}
\end{table*}

\noindent \textbf{Colour Evolution Stage.} After the distillation of a certain colour naming system of human language, we evolve the CQFormer to split  fine-grained new colour. In this stage, we remove the restriction on the number of quantised colours and minimise different combinations of losses to encourage the CQFormer to evolve more colours.  Please refer to  Sec.~\ref{sec:Colour Evolution Evaluation} for more details.

\section{Experiments}

We evaluate the CQFormer on mainstream benchmark datasets of both object detection task (Sec.~\ref{sec:det}) and image classification task (Sec.~\ref{sec:cls}). Additionally, we specifically design a colour evolution experiment (Sec.~\ref{sec:Colour Evolution Evaluation}) to demonstrate how our CQFormer automatically evolves to increase fine-grained colours. For ablation study, visualisation and other details, please refer to Supplementary Material.

\subsection{Datasets and Experiment Settings}
\noindent \textbf{Datasets.} For object detection, we utilise MS COCO~\cite{COCO} dataset, which contains $\sim$118k images with bounding box annotation in 80 categories. Here, we use COCO~\texttt{train2017} set as the train set and use COCO~\texttt{val2017} set as the test set. For classification, we use CIFAR10, CIFAR100~\cite{cifar10}, STL10~\cite{stl10} and tiny-imagenet-200 (Tiny200)~\cite{le2015tiny} dataset. Both CIFAR10 and CIFAR100 contain 60000 images, covering 10 and 100 classes of general objects, respectively. The STL10 consists of 13000 images (5000 for training and 8000 for testing) with 10 classes. The Tiny200 is a subset of ImageNet~\cite{imagenet_cvpr09} and contains 200 categories with 110k images. We adopt random crop and  random flip for data augmentation.

\noindent \textbf{Evaluation Metrics.} For object detection, we report average precision (AP) value. For classification, we report top-1 classification accuracy.


\noindent \textbf{Implement Details.}

\noindent \textit{(1) Upper bound.} We utilise the performance of the detector/classifier without additional colour quantisation methods in full-colour space (24-bit) as the upper bound.  For object detection, we adopt the Sparse-RCNN~\cite{sun2021sparse} detector with ResNet-50~\cite{he2016deep} backbone and FPN~\cite{fpn} neck. For classification, we adopt Resnet-18~\cite{he2016deep} network. 

\noindent \textit{(2) Training Settings.} All colour quantisation experiments are finished at quantisation levels from 1-bit to 6-bit, \textit{i.e.} $C \in \{2,4,8,16,32,64\}$. We set the temperature parameter $\tau = 0.01$ and the $\alpha = 1$, $\beta = 0.3$ , $\gamma = 1$. For detection, only the Resnet-50 backbone of the detector is initialised with Imagenet \cite{imagenet_cvpr09} pre-train weights.  We jointly train our CQFormer with Sparse-RCNN for 36 epochs on 4 Tesla V100 GPUs. The batch size is set to 8, and the optimiser is  AdamW \cite{loshchilov2018decoupled}. The initial learning rate is set to $1.25e^{-6}$, and the learning rate would decay to one-tenth at 24 epochs. For classification, we cascade CQFormer with the Resnet-18 and jointly train them  without pre-trained models on a single GeForce RTX A6000 GPU.  We employ an SGD optimiser for 60 epochs using the Cosine-Warm-Restart~\cite{loshchilov2016sgdr} as the learning rate scheduler. A batch size of 128 (STL10 is set to 32), an initial learning rate of 0.05, a momentum of 0.5, and a weight decay of 0.001 are used.

\noindent \textbf{Comparison Methods.} We compare with three traditional perception-centred methods: MedianCut~\cite{mediancut} , MedianCut+Dither~\cite{mediancut_with_dither} and OCTree~\cite{octree}, and a task-centred CNN-based method ColorCNN~\cite{hou2020learning}. Specifically, for ColorCNN, we adopt the same training strategy as we did for CQFormer. For the traditional methods, we conduct comparative experiments as described in \cite{hou2020learning}.



\subsection{Object Detection Task Evaluation}
\label{sec:det}
Table.~\ref{Tab:Detection_sparse} shows the object detection results on MS COCO dataset~\cite{COCO} using Sparse-RCNN~\cite{sun2021sparse} detector. Our CQFormer outperforms all other methods in terms of AP value performance under all colour quantisation levels, from 1-bit to 6-bit. This substantial improvement demonstrates the effectiveness of our CQFormer in the object detection task. We also evaluate CQFormer with other popular detectors in Supplementary Material.

\subsection{Classification Task Evaluation}
\label{sec:cls}

\begin{figure*}[ht!]
    \centering
    \includegraphics[width=1.0\linewidth]{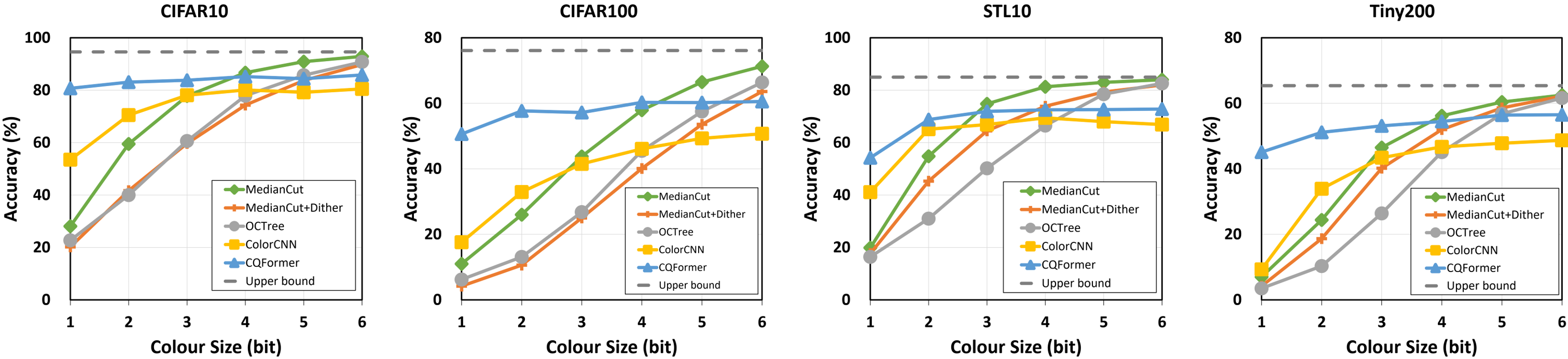}
    \vspace{-0.5cm}
    \caption{Top-1 classification accuracy of colour-quantised images on four datasets with Resnet-18~\cite{he2016deep} networks.}
    \label{exp:cls}
    \vspace{-0.1cm}
\end{figure*}

\begin{figure*}[ht!]
 \centering
 \includegraphics[width=1.0\linewidth]{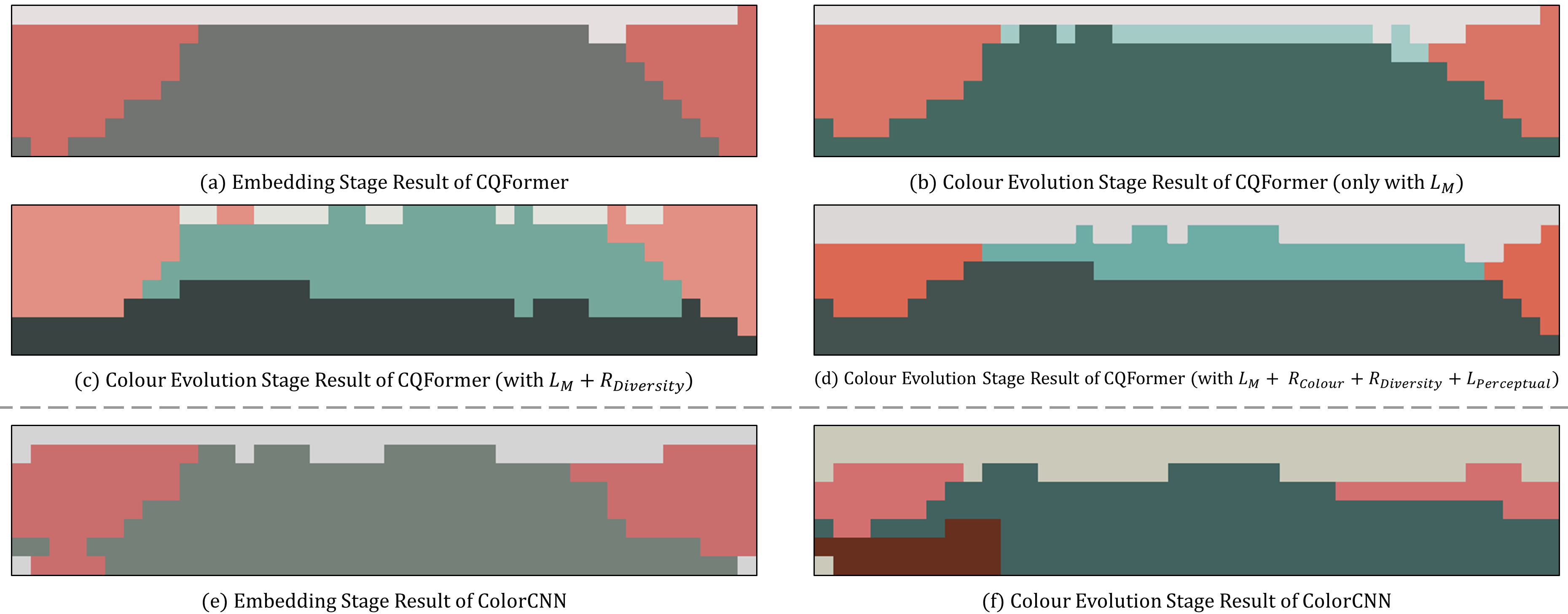}
  \vspace{-7mm}
 \caption{(a) is the embedding stage result of CQFormer. (b) (c) (d) are the colour evolution stage results of CQFormer with different combinations. (e)  is the embedding stage result of ColorCNN \cite{hou2020learning}. (f) is the colour evolution stage result of ColorCNN \cite{hou2020learning}.}
 \label{fig:Results of Colour Evolution Evaluation}
 \vspace{-0.3cm}
\end{figure*}

\noindent \textbf{Results.} Fig.~\ref{exp:cls} presents comparisons to the state-of-the-art methods on the four datasets. Our proposed CQFormer (solid blue line) has a consistent and obvious improvement over all other methods in extremely low-bit colour space (less than 3-bit). Moreover, our CQFormer archives a superior performance than the task-centred method ColorCNN~\cite{hou2020learning} under all colour quantisation levels from 1-bit to 6-bit.

\noindent \textbf{Discussion of classification under a large colour space.} Similar to the task-centred ColorCNN, our classification performance is also inferior to the traditional method under a large colour space (greater than 4-bit), despite superior performance on object detection task. As extensively discussed in~\cite{hou2020learning}, this is a common characteristic for non-clustering based quantisation. 
 
Here we are satisfied with our classification performance under the small colour space for two reasons: First of all,  most human languages only use fewer than 3-bit colour terms. This implies that discovering more colours not only compromises the principle of efficiency but also goes contrary to the expectation of better perceptual effects. In other words, the performance of the CQFormer on limited colour categories may hint at the optimal outcome restricted by the unique neurological structure of human vision and cognition, which are reflected in a wide array of languages in turn. 

Secondly, larger colour space naturally preserves more visual fidelity. 6-bit colour space could already deliver a vivid image, and the recognition performance on these images is only marginally below the original image. It could neither save much storage space nor reveal special knowledge. Therefore, we only focus on the small colour space.




\subsection{Colour Evolution Evaluation}
\label{sec:Colour Evolution Evaluation}

\noindent \textbf{Settings.} This experiment is based on the classification task on CIFAR10~\cite{cifar10} dataset using the Resnet-18~\cite{he2016deep} classifier. In the embedding stage, we embed the 1978 Nafaanra three-colour system (Fig.~\ref{fig:inject_human_language} (c)) using the colour probability map embedding and set $C$ as 3. The reason why we choose Nafaanra is analysed in the Supplementary Material. Here we set $\tau = 1.0$, and force the fourth colour to be split from light (`fiNge'), dark (`wOO'), and warm or red-like (`nyiE'), respectively. We optimise the loss function $\mathcal{L_{\operatorname{Full-Embedding}}}$ in Eq.~\ref{eq:loss1} for the initial 40 epochs.  In the colour evolution stage, we design various combinations of loss function, \textit{i.e.} only $\mathcal{L}_{M}$,  $\mathcal{L}_{M}+R_{\operatorname{Diversity}}$ and $\mathcal{L}_{M}+R_{\operatorname{Colour}}+R_{\operatorname{Diversity}}+\mathcal{L}_{\operatorname{Perceptual}}$, and minimise them separately for the subsequent 20 epochs without the restriction of colour size. Additionally, we also conduct a colour evolution experiment on the \textit{task-centred} ColorCNN \cite{hou2020learning} for comparison, which is slightly modified to implement the same embedding and evolution details.


\noindent \textbf{Results and Analysis.} The results of colour evolution evaluation is shown in Fig. \ref{fig:Results of Colour Evolution Evaluation}. First, during the embedding stage, the WCS colour probability map generated by the  CQFormer (Fig. \ref{fig:Results of Colour Evolution Evaluation} (a)) is similar to Nafaanra (Fig.\ref{fig:inject_human_language} (c)). Therefore, we successfully embed the colour naming system of human language into the latent space of CQFormer. Second, during the colour evolution stage, as shown in Fig. \ref{fig:Results of Colour Evolution Evaluation} (b)(c)(d), the CQFormer automatically evolves the fourth colour that is split from dark (`wOO') and close to yellow-green under \textit{all combinations}, matching the basic colour terms theory~\cite{kay1978linguistic}. Third, although the ColorCNN also evolves the fourth colour close to brown (Fig. \ref{fig:Results of Colour Evolution Evaluation} (f)), it fails to match the basic colour terms theory~\cite{kay1978linguistic} since the yellow-green and blue are skipped. 

We are not able to see the fourth colour split from either light (`fiNge') or warm/red-like (`nyiE') in the WCS colour probability map, since only 3.7\% (if split from light (`fiNge')) or 5.9\% (if split from  warm/red-like (`nyiE')) of all pixels are assigned to the fourth colour, compared with 23.5\% (split from dark (`wOO')). Very interestingly, this phenomenon echoes the evolution of the information bottleneck (IB) colour naming systems \cite{zaslavsky2018efficient}, where the fourth colour should be spilt from dark in the "dark-light-red" colour scheme.

Although the CQFormer can also evolve the fourth colour automatically when optimising $\mathcal{L}_{M}$ alone, it only covers a small portion of the WCS probability map in Fig. \ref{fig:Results of Colour Evolution Evaluation} (b). As the items of the perceptual structure loss increase, \textit{i.e.} from Fig. \ref{fig:Results of Colour Evolution Evaluation} (b) to (d), these four colours have clearer borders, a more logical proportion and are each more internally clustered. This suggests that the colour evolution is not fully complete when considering machine accuracy alone. In other words, thanks to the CQFormer's ability to integrate the need for both machine accuracy and human perception, the discovered colour naming system evolves more thoroughly and effectively, which is mirrored in colour naming patterns of human language~\cite{Zaslavsky2022evolution}.



\section{Limitation and Discussion}



While the complexity-accuracy trade-off of machine-discovered colour terms, as shown in Fig.~\ref{fig:overview} (b), is quite similar to the numerical limit of categorical counterparts for human languages, the current work is still preliminary. As shown in Fig.~\ref{fig:overview}, the newly discovered WCS colour probability map is still quite different from the human one. A more accurate language evolution replication needs to consider more complex variables such as environmental idiosyncrasies, cultural peculiarities, functional necessities, technological maturity, learning experience, and intercultural communication.  

Another promising direction would be associating the discovered colours with human colour terms. This would involve much research on Nature Language Processing, and we hope to discuss it with experts from different disciplines in future works. Last but not least, the AI simulation outcome contributes to the long-standing universalist-relativist debate of the linguistic community on colour cognition. Though not entirely excluding the cultural specificity of the colour schemes, the machine finding strongly supports the universalist view that an innate, physiological principle constraints, if not determine, the evolutionary sequence and distributional possibilities of basic colour terms in communities of different cultural traditions. The complexity-efficiency principle is confirmed by the finding that the numerical limitation of colour categories could lead to superior performance on colour-specific tasks, contrary to the intuitive expectation that complexity breeds perfection. The independent AI discovery of the "green-yellow" category on the basis of the fundamental tripartite "dark-light-red" colour scheme points to the congruence of neural algorithms and human cognition and opens a new frontier to test contested hypothesis in the social sciences through machine simulation. We would be more than delighted if this tentative attempt would prove to be a bridge to link scholars of different disciplines for more collaboration and generate more fruitful results.

\section*{Acknowledgement}
This work was supported by JST Moonshot R\&D Grant Number JPMJMS2011 and JST ACT-X Grant Number JPMJAX190D, Japan.

{\small
\bibliographystyle{ieee_fullname}
\bibliography{egbib}

\begin{thebibliography}{10}\itemsep=-1pt

\bibitem{bahng2018coloring}
Hyojin Bahng, Seungjoo Yoo, Wonwoong Cho, David~Keetae Park, Ziming Wu,
  Xiaojuan Ma, and Jaegul Choo.
\newblock Coloring with words: Guiding image colorization through text-based
  palette generation.
\newblock In {\em Proceedings of the european conference on computer vision
  (eccv)}, pages 431--447, 2018.

\bibitem{berlin1969basic}
B. Berlin and P. Kay.
\newblock {\em Basic Color Terms: Their Universality and Evolution}.
\newblock University of California Press, 1969.

\bibitem{boutell1997png}
Thomas Boutell.
\newblock Png (portable network graphics) specification version 1.0.
\newblock Technical report, 1997.

\bibitem{camposeco2019hybrid}
Federico Camposeco, Andrea Cohen, Marc Pollefeys, and Torsten Sattler.
\newblock Hybrid scene compression for visual localization.
\newblock In {\em Proceedings of the IEEE/CVF Conference on Computer Vision and
  Pattern Recognition}, pages 7653--7662, 2019.

\bibitem{carion2020detr}
Nicolas Carion, Francisco Massa, Gabriel Synnaeve, Nicolas Usunier, Alexander
  Kirillov, and Sergey Zagoruyko.
\newblock End-to-end object detection with transformers.
\newblock In {\em European conference on computer vision}, pages 213--229.
  Springer, 2020.

\bibitem{Communicating_pnas}
Rahma Chaabouni, Eugene Kharitonov, Emmanuel Dupoux, and Marco Baroni.
\newblock Communicating artificial neural networks develop efficient
  color-naming systems.
\newblock {\em Proceedings of the National Academy of Sciences},
  118(12):e2016569118, 2021.

\bibitem{chang2015palette}
Huiwen Chang, Ohad Fried, Yiming Liu, Stephen DiVerdi, and Adam Finkelstein.
\newblock Palette-based photo recoloring.
\newblock {\em ACM Trans. Graph.}, 34(4):139--1, 2015.

\bibitem{cho2017palettenet}
Junho Cho, Sangdoo Yun, Kyoung Mu~Lee, and Jin Young~Choi.
\newblock Palettenet: Image recolorization with given color palette.
\newblock In {\em Proceedings of the ieee conference on computer vision and
  pattern recognition workshops}, pages 62--70, 2017.

\bibitem{chollet2017xception}
Fran{\c{c}}ois Chollet.
\newblock Xception: Deep learning with depthwise separable convolutions.
\newblock In {\em Proceedings of the IEEE conference on computer vision and
  pattern recognition}, pages 1251--1258, 2017.

\bibitem{stl10}
Adam Coates, Andrew Ng, and Honglak Lee.
\newblock An analysis of single-layer networks in unsupervised feature
  learning.
\newblock In {\em Proceedings of the fourteenth international conference on
  artificial intelligence and statistics}, pages 215--223. JMLR Workshop and
  Conference Proceedings, 2011.

\bibitem{Cui_2022_BMVC}
Ziteng Cui, Kunchang Li, Lin Gu, Shenghan Su, Peng Gao, ZhengKai Jiang, Yu
  Qiao, and Tatsuya Harada.
\newblock You only need 90k parameters to adapt light: a light weight
  transformer for image enhancement and exposure correction.
\newblock In {\em 33rd British Machine Vision Conference 2022, {BMVC} 2022,
  London, UK, November 21-24, 2022}. {BMVA} Press, 2022.

\bibitem{Emergent_eLife}
Jelmer~P de Vries, Arash Akbarinia, Alban Flachot, and Karl~R Gegenfurtner.
\newblock Emergent color categorization in a neural network trained for object
  recognition.
\newblock {\em Elife}, 11:e76472, 2022.

\bibitem{imagenet_cvpr09}
J. Deng, W. Dong, R. Socher, L.-J. Li, K. Li, and L. Fei-Fei.
\newblock {ImageNet: A Large-Scale Hierarchical Image Database}.
\newblock In {\em CVPR09}, 2009.

\bibitem{deng1999peer}
Yining Deng, Charles Kenney, Michael~S Moore, and BS Manjunath.
\newblock Peer group filtering and perceptual color image quantization.
\newblock In {\em 1999 IEEE International Symposium on Circuits and Systems
  (ISCAS)}, volume~4, pages 21--24. IEEE, 1999.

\bibitem{dosovitskiy2020vit}
Alexey Dosovitskiy, Lucas Beyer, Alexander Kolesnikov, Dirk Weissenborn,
  Xiaohua Zhai, Thomas Unterthiner, Mostafa Dehghani, Matthias Minderer, Georg
  Heigold, Sylvain Gelly, Jakob Uszkoreit, and Neil Houlsby.
\newblock An image is worth 16x16 words: Transformers for image recognition at
  scale.
\newblock {\em ICLR}, 2021.

\bibitem{dithering}
R.~W. Floyd and L. Steinberg.
\newblock An adaptive algorithm for spatial gray scale.
\newblock {\em Proceedings of the Society for Information Display}, 17, 1975.

\bibitem{mediancut_with_dither}
R.~W. Floyd and L. Steinberg.
\newblock An adaptive algorithm for spatial grayscale.
\newblock {\em Proceedings of the Society for Information Display}, 17, 1976.

\bibitem{forsyth2002computer}
David~A Forsyth and Jean Ponce.
\newblock {\em Computer vision: a modern approach}.
\newblock prentice hall professional technical reference, 2002.

\bibitem{octree}
Michael Gervautz and Werner Purgathofer.
\newblock A simple method for color quantization: Octree quantization.
\newblock In {\em New trends in computer graphics}, pages 219--231. Springer,
  1988.

\bibitem{gibson2017color}
Edward Gibson, Richard Futrell, Julian Jara-Ettinger, Kyle Mahowald, Leon
  Bergen, Sivalogeswaran Ratnasingam, Mitchell Gibson, Steven~T Piantadosi, and
  Bevil~R Conway.
\newblock Color naming across languages reflects color use.
\newblock {\em Proceedings of the National Academy of Sciences},
  114(40):10785--10790, 2017.

\bibitem{he2016deep}
Kaiming He, Xiangyu Zhang, Shaoqing Ren, and Jian Sun.
\newblock Deep residual learning for image recognition.
\newblock In {\em Proceedings of the IEEE conference on computer vision and
  pattern recognition}, pages 770--778, 2016.

\bibitem{softmax_temperature}
Yu-Lin He, Xiao-Liang Zhang, Wei Ao, and Joshua~Zhexue Huang.
\newblock Determining the optimal temperature parameter for softmax function in
  reinforcement learning.
\newblock {\em Applied Soft Computing}, 70:80--85, 2018.

\bibitem{mediancut}
Paul Heckbert.
\newblock Color image quantization for frame buffer display.
\newblock {\em ACM Siggraph Computer Graphics}, 16(3):297--307, 1982.

\bibitem{hendrycks2016gaussian}
Dan Hendrycks and Kevin Gimpel.
\newblock Gaussian error linear units (gelus).
\newblock {\em arXiv preprint arXiv:1606.08415}, 2016.

\bibitem{hou2020learning}
Yunzhong Hou, Liang Zheng, and Stephen Gould.
\newblock Learning to structure an image with few colors.
\newblock In {\em Proceedings of the IEEE/CVF Conference on Computer Vision and
  Pattern Recognition}, pages 10116--10125, 2020.

\bibitem{kay2009world}
Paul Kay, Brent Berlin, Luisa Maffi, William~R Merrifield, and Richard Cook.
\newblock {\em The world color survey}.
\newblock CSLI Publications Stanford, CA, 2009.

\bibitem{kay1978linguistic}
Paul Kay and Chad~K McDaniel.
\newblock The linguistic significance of the meanings of basic color terms.
\newblock {\em Language}, 54(3):610--646, 1978.

\bibitem{cifar10}
Alex Krizhevsky, Geoffrey Hinton, et~al.
\newblock Learning multiple layers of features from tiny images.
\newblock 2009.

\bibitem{le2015tiny}
Ya Le and Xuan Yang.
\newblock Tiny imagenet visual recognition challenge.
\newblock {\em CS 231N}, 7(7):3, 2015.

\bibitem{li2022neural}
Boyi Li, Serge Belongie, Ser-nam Lim, and Abe Davis.
\newblock Neural image recolorization for creative domains.
\newblock In {\em Proceedings of the IEEE/CVF Conference on Computer Vision and
  Pattern Recognition}, pages 2226--2230, 2022.

\bibitem{li2021pose}
Ke Li, Shijie Wang, Xiang Zhang, Yifan Xu, Weijian Xu, and Zhuowen Tu.
\newblock Pose recognition with cascade transformers.
\newblock In {\em Proceedings of the IEEE/CVF Conference on Computer Vision and
  Pattern Recognition}, pages 1944--1953, 2021.

\bibitem{liang2003general}
DianLong Liang, Lizhi Cheng, and Zenghui Zhang.
\newblock General construction of wavelet filters via a lifting scheme and its
  application in image coding.
\newblock {\em Optical Engineering}, 42(7):1949--1955, 2003.

\bibitem{fpn}
Tsung-Yi Lin, Piotr Doll{\'a}r, Ross Girshick, Kaiming He, Bharath Hariharan,
  and Serge Belongie.
\newblock Feature pyramid networks for object detection.
\newblock In {\em Proceedings of the IEEE conference on computer vision and
  pattern recognition}, pages 2117--2125, 2017.

\bibitem{COCO}
Tsung-Yi Lin, Michael Maire, Serge Belongie, James Hays, Pietro Perona, Deva
  Ramanan, Piotr Doll{\'a}r, and C.~Lawrence Zitnick.
\newblock Microsoft coco: Common objects in context.
\newblock In David Fleet, Tomas Pajdla, Bernt Schiele, and Tinne Tuytelaars,
  editors, {\em Computer Vision -- ECCV 2014}, pages 740--755, Cham, 2014.
  Springer International Publishing.

\bibitem{loshchilov2016sgdr}
Ilya Loshchilov and Frank Hutter.
\newblock Sgdr: Stochastic gradient descent with warm restarts.
\newblock {\em arXiv preprint arXiv:1608.03983}, 2016.

\bibitem{loshchilov2018decoupled}
Ilya Loshchilov and Frank Hutter.
\newblock Decoupled weight decay regularization.
\newblock In {\em International Conference on Learning Representations}, 2019.

\bibitem{CQ_of_images}
M.T. Orchard and C.A. Bouman.
\newblock Color quantization of images.
\newblock {\em IEEE Transactions on Signal Processing}, 39(12):2677--2690,
  1991.

\bibitem{NICE_PLOS}
C~Alejandro Parraga and Arash Akbarinia.
\newblock Nice: A computational solution to close the gap from colour
  perception to colour categorization.
\newblock {\em PloS one}, 11(3):e0149538, 2016.

\bibitem{poynton2012digital}
Charles Poynton.
\newblock {\em Digital video and HD: Algorithms and Interfaces}.
\newblock Elsevier, 2012.

\bibitem{MohammadECCV16binary}
Mohammad Rastegari, Vicente Ordonez, Joseph Redmon, and Ali Farhadi.
\newblock Xnor-net: Imagenet classification using binary convolutional neural
  networks.
\newblock In Bastian Leibe, Jiri Matas, Nicu Sebe, and Max Welling, editors,
  {\em Computer Vision -- ECCV 2016}, pages 525--542, Cham, 2016. Springer
  International Publishing.

\bibitem{shannon1948mathematical}
Claude~Elwood Shannon.
\newblock A mathematical theory of communication.
\newblock {\em The Bell system technical journal}, 27(3):379--423, 1948.

\bibitem{siuda2019color}
Katarzyna Siuda-Krzywicka, Christoph Witzel, Emma Chabani, Myriam Taga, Cecile
  Coste, Noella Cools, Sophie Ferrieux, Laurent Cohen, Tal~Seidel Malkinson,
  and Paolo Bartolomeo.
\newblock Color categorization independent of color naming.
\newblock {\em Cell reports}, 28(10):2471--2479, 2019.

\bibitem{sun2021sparse}
Peize Sun, Rufeng Zhang, Yi Jiang, Tao Kong, Chenfeng Xu, Wei Zhan, Masayoshi
  Tomizuka, Lei Li, Zehuan Yuan, Changhu Wang, and Ping Luo.
\newblock Sparse r-cnn: End-to-end object detection with learnable proposals.
\newblock In {\em Proceedings of the IEEE/CVF conference on computer vision and
  pattern recognition}, pages 14454--14463, 2021.

\bibitem{valanarasu2022unext}
Jeya Maria~Jose Valanarasu and Vishal~M Patel.
\newblock Unext: Mlp-based rapid medical image segmentation network.
\newblock {\em arXiv preprint arXiv:2203.04967}, 2022.

\bibitem{van2007learning}
Joost Van De~Weijer, Cordelia Schmid, and Jakob Verbeek.
\newblock Learning color names from real-world images.
\newblock In {\em 2007 IEEE conference on computer vision and pattern
  recognition}, pages 1--8. IEEE, 2007.

\bibitem{wu1992color}
Xiaolin Wu.
\newblock Color quantization by dynamic programming and principal analysis.
\newblock {\em ACM Transactions on Graphics (TOG)}, 11(4):348--372, 1992.

\bibitem{yang2019quantization}
Jiwei Yang, Xu Shen, Jun Xing, Xinmei Tian, Houqiang Li, Bing Deng, Jianqiang
  Huang, and Xian-sheng Hua.
\newblock Quantization networks.
\newblock In {\em Proceedings of the IEEE/CVF Conference on Computer Vision and
  Pattern Recognition}, pages 7308--7316, 2019.

\bibitem{yoo2019coloring}
Seungjoo Yoo, Hyojin Bahng, Sunghyo Chung, Junsoo Lee, Jaehyuk Chang, and
  Jaegul Choo.
\newblock Coloring with limited data: Few-shot colorization via memory
  augmented networks.
\newblock In {\em Proceedings of the IEEE/CVF Conference on Computer Vision and
  Pattern Recognition}, pages 11283--11292, 2019.

\bibitem{Zaslavsky2022evolution}
Noga Zaslavsky, Karee Garvin, Charles Kemp, Naftali Tishby, and Terry Regier.
\newblock {The evolution of color naming reflects pressure for efficiency:
  {E}vidence from the recent past}.
\newblock {\em Journal of Language Evolution}, 2022.

\bibitem{zaslavsky2018efficient}
Noga Zaslavsky, Charles Kemp, Terry Regier, and Naftali Tishby.
\newblock Efficient compression in color naming and its evolution.
\newblock {\em Proceedings of the National Academy of Sciences},
  115(31):7937--7942, 2018.

\bibitem{Zaslavsky19a}
Noga Zaslavsky, Charles Kemp, Naftali Tishby, and Terry Regier.
\newblock Color naming reflects both perceptual structure and communicative
  need.
\newblock {\em Topics in Cognitive Science}, 11(1):207--219, 2019.

\bibitem{Zaslavsky2019b}
Noga Zaslavsky, Charles Kemp, Naftali Tishby, and Terry Regier.
\newblock Communicative need in color naming.
\newblock {\em Cognitive Neuropsychology}, 2019.

\end{thebibliography}
}

\end{document}


\title{Name Your Colour For the Task: Artificially Discover Colour Naming \\ via  Colour Quantisation Transformer \\ Supplementary Material}

\author{Shenghan Su\textsuperscript{1},
    Lin Gu\textsuperscript{3,2}\thanks{Corresponding author.} \:,
    Yue Yang\textsuperscript{1,4},
    Zenghui Zhang\textsuperscript{1},
    Tatsuya Harada\textsuperscript{2,3}\\
    \textsuperscript{1}Shanghai Jiao Tong University, \textsuperscript{2}The University of Tokyo,  
    \textsuperscript {3}RIKEN AIP,
    \textsuperscript {4} Shanghai AI Laboratory\\
    \footnotesize{\texttt{\{su2564468850, yang-yue, zenghui.zhang\}@sjtu.edu.cn, lin.gu@riken.jp, harada@mi.t.u-tokyo.ac.jp}}
    }

\maketitle
\ificcvfinal\thispagestyle{empty}\fi


\appendix

\section{Ablation Study}
\label{sec:Ablation}

\begin{table*}[htp!]
\renewcommand\arraystretch{1.2}
\centering
\caption{Ablation study results under 1-bit colour quantisation, Tiny200~\cite{le2015tiny} dataset, Renset18~\cite{he2016deep} classifier.}
\label{tab:ablation}
\vspace{2mm}
\small
\begin{tabular}{ccccc|c}
\hline
$\tau$ & $R_{\operatorname{Colour}}$  & $R_{\operatorname{Diversity}}$  & $\mathcal{L}_{\operatorname{Perceptual}}$ & Palette Branch  & Accuracy \\ \hline
\CheckmarkBold & \CheckmarkBold & \CheckmarkBold & \CheckmarkBold& \CheckmarkBold & \textbf{45.1}          \\ \hline
\XSolidBrush & \color{lightgray}{\CheckmarkBold} & \color{lightgray}{\CheckmarkBold} & \color{lightgray}{\CheckmarkBold}& \color{lightgray}{\CheckmarkBold} & 6.9           \\
\color{lightgray}{\CheckmarkBold} & \XSolidBrush & \color{lightgray}{\CheckmarkBold} & \color{lightgray}{\CheckmarkBold}& \color{lightgray}{\CheckmarkBold} & 43.9          \\
\color{lightgray}{\CheckmarkBold} & \color{lightgray}{\CheckmarkBold} & \XSolidBrush & \color{lightgray}{\CheckmarkBold} & \color{lightgray}{\CheckmarkBold} & 43.7          \\
\color{lightgray}{\CheckmarkBold} & \color{lightgray}{\CheckmarkBold} & \color{lightgray}{\CheckmarkBold} & \XSolidBrush & \color{lightgray}{\CheckmarkBold} & 44.6  \\
\color{lightgray}{\CheckmarkBold} & \XSolidBrush & \XSolidBrush & \XSolidBrush & \color{lightgray}{\CheckmarkBold}& 37.3  \\ 
\color{lightgray}{\CheckmarkBold} & \color{lightgray}{\CheckmarkBold} & \color{lightgray}{\CheckmarkBold} & \color{lightgray}{\CheckmarkBold} &  \XSolidBrush& 30.4  \\ \hline
\end{tabular}
\end{table*}

\begin{table*}[htp!]
\renewcommand\arraystretch{1.2}
\centering
\caption{Ablation study results of robustness under under different colour quantisation levels from 1-bit to 4-bit, CIFAR10~\cite{cifar10} dataset, Renset18~\cite{he2016deep} classifier.}
\label{tab:ablation2}
\vspace{2mm}
\begin{tabular}{l|cccc}
\hline
                           & 1 bit & 2 bit & 3 bit & 4 bit \\ \hline
CQFormer                   & \textbf{80.7}  & \textbf{83.1}  & \textbf{83.8}  & \textbf{85.2}  \\ \hline
CQFormer (with colour jitter) & 79.3  & 81.6  & 83.4  & 84.6  \\
CQFormer(with Gaussian blur )           & 80.6  & 81.7  & 81.9  & 83.6  \\ \hline
\end{tabular}
\end{table*}

 \begin{figure}[h!]
 \centering

 \includegraphics[width=\linewidth]{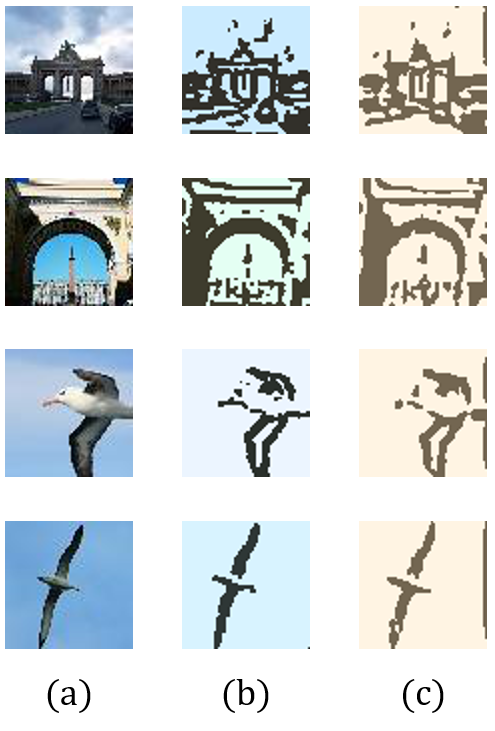}

 \caption{(a) is the original image. (b) is the quantised image with Palette Branch. (c) is the quantised image using a set of centroids instead of Palette Branch.}
 \label{fig:ablation}
\end{figure}

As shown in Table \ref{tab:ablation}, we ablate the important elements in our CQFormer, using Tiny200~\cite{le2015tiny} classification dataset  under 1-bit colour quantisation. We investigate the effectiveness of the temperature parameter by setting $\tau = 1.0$, perceptual structure loss by setting any terms of  $\alpha$, $\beta$ and $\gamma$  as 0 and Palette Branch by replacing it with a set of centroids in colour space. As shown in Table \ref{tab:ablation2}, we also investigate the robustness of CQFormer by adding colour jitter and Gaussian blur using CIFAR10~\cite{cifar10} classification dataset under different colour quantisation levels from 1-bit to 4-bit.

\noindent \textbf{Influence of temperature parameter:} Without the temperature parameter, a severe  accuracy drop (-38.8\%) has occurred, which shows that the temperature parameter in our CQFormer can further approximate the $\operatorname{One-hot}(M(\boldsymbol{x}))$ using $m_{\tau}(\boldsymbol{x})$ during training stage to boost classification accuracy.

\noindent \textbf{Influence of perceptual structure loss:} With the  $R_{\operatorname{Colour}}$, $R_{\operatorname{Diversity}}$ and $\mathcal{L}_{\operatorname{Perceptual}}$, our CQFormer improve the top-1 accuracy by 1.2\%, 1.4\% and 0.5\%, respectively. When we remove all of them, a considerable accuracy drop of -7.8\% has occurred. It demonstrates that the perceptual structure loss contributes to better machine accuracy besides maintaining perceptual similarity.

\noindent \textbf{Influence of Palette Branch:}  If we remove Palette Branch and make the CQFormer just learn a set of centroids in RGB colour space, the accuracy is 30.4\%, resulting in a severe drop of 14.7\%. As shown in Fig. \ref{fig:ablation}, it would make all the images have the same colour palette rather than the same amount of colours, resulting in a loss of perceptual similarity, \textit{e.g.} the blue sky in Col.(c) is represented as light yellow. Therefore, our Palette Branch ensures  that reference palette queries are sent to interact with the image and creates the colour palette using both machine preference and image perception features, which maintains the colour specificity of each image.

\noindent \textbf{Robustness of CQFormer: } We add a colour jitter and Gaussian blur to the colour-quantised image. Results are shown in Table. \ref{tab:ablation2}. For example, on the CIFAR10 classification dataset, we achieve 79.3\%, 81.6\%, 83.4\%, and 84.6\% top-1 accuracy with a colour jitter from 1-bit to 4-bit colour quantisation. The colour jitter causes a little drop of 1.4\%, 1.5\%, 0.4\%, and 0.6\%, respectively. Therefore, our CQFormer is robust enough to overcome different noises.

\section{Object Detection with other detector}
\label{Appendix:other detector}

\vspace{-0.2cm}
\begin{table*}[htp]
\centering
\renewcommand\arraystretch{1.1}
\caption{Object detection results on MS COCO dataset~\cite{COCO} with Faster-RCNN~\cite{ren2015faster} detector, here we report the average precision (AP) value.}
\vspace{2mm}
\begin{adjustbox}{max width = \linewidth}
\begin{tabular}{l|ccccccc}
\hline
Method             & 1-bit & 2-bit & 3-bit & 4-bit & 5-bit & 6-bit & Full Colour (24-bit) \\ \hline
Upper bound        & -     & -     & -     & -     & -     & -     & 37.2          \\ \hline
Median Cut (w/o D)~\cite{mediancut} & 6.5   & 9.7   & 11.4  & 14.0  & \textbf{16.8}  & 18.1  & -             \\ \hline
Median Cut (w D)~\cite{mediancut_with_dither}   & 7.3   & 8.8  & 11.2  & 13.4  & 14.8  & 15.6  & -             \\ \hline
OCTree~\cite{octree}             & 8.7   & 9.0   & 9.8   & 10.9  & 13.4  & 14.5  & -             \\ \hline
\textbf{CQFormer}           & \textbf{8.9}  & \textbf{11.2}  & \textbf{12.8}  & \textbf{14.5}  & 16.6  & \textbf{19.4}  & -             \\ \hline
\end{tabular}
\end{adjustbox}
\label{Tab:Detection}
\end{table*}

To evaluate our CQFormer's generalisation on another object detector, we use the popular object detector Faster-RCNN~\cite{ren2015faster} for evaluation. We jointly train our CQFormer with Faster-RCNN for 12 epochs on 4 Tesla V100 GPUs.  We adopt the SGD optimiser with an initial learning rate $0.01$, momentum $0.9$ and weight decay $0.0001$, learning rate decay to one-tenth at 8 and 11 epochs. The dataset and other training settings are the same as  Sparse-RCNN~\cite{sun2021sparse} experiments in Sec. 4.2.


\section{The Reason of Choosing Nafaanra for Colour Evolution Experiments}
\label{Appendix:Explanation of Nafaanra Choice}
Colour naming data for Nafaanra \cite{Zaslavsky2022evolution} were initially collected in 1978 in Banda Ahenkro, as part of the WCS \cite{berlin1969basic}, which is a 3-term system, with terms for light (`fiNge'), dark (`wOO'), and warm or red-like (`nyiE'). In 2018, 40 years after the original WCS data collection, Nafaanra colour naming data were collected again in the same town. As technology and lifestyle changed in the community, the Nafaanra colour naming system changed substantially between 1978 and 2018, becoming more semantically fine-grained by adding new colour terms and adjusting the extension of previously existing terms \cite{Zaslavsky2022evolution}. Therefore, the Nafaanra language is an excellent example for comparing  the evolution trajectory of human language using the machine.

\section{The Relationship between the IB Color Naming Model and CQFormer}
\begin{figure*}[h!]
 \centering
 \includegraphics[width=1.0\linewidth]{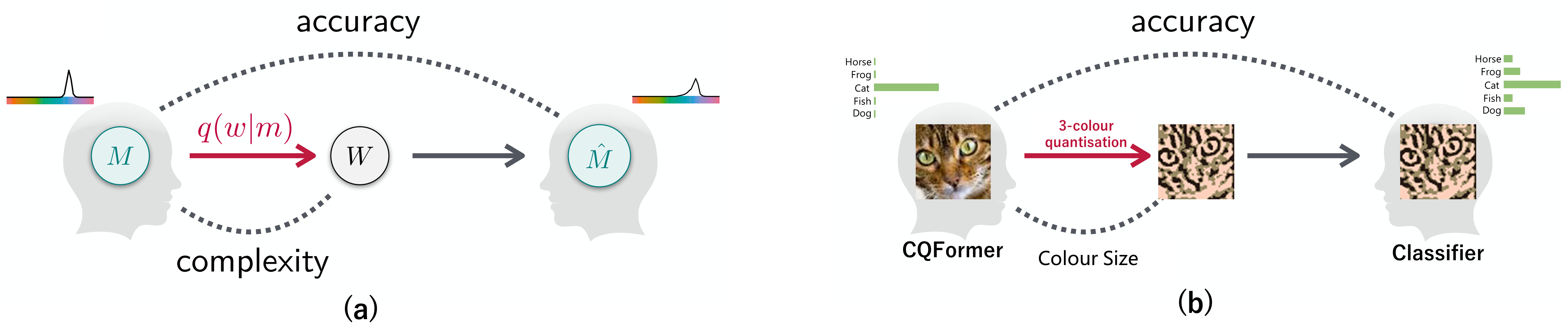}
 \caption{(a) is the IB color naming model proposed by \cite{zaslavsky2018efficient}.(b) is our colour quantisation model.}
 \label{fig:IB_model}
 \vspace{-0.2cm}
\end{figure*}

Fig. \ref{fig:IB_model}(a) is the information bottleneck (IB) color naming model proposed by \cite{zaslavsky2018efficient}. A straightforward communication scenario, which can be derived from Shannon's communication model \cite{shannon1948mathematical}, serves as the foundation for this theoretical framework.

Fig. \ref{fig:IB_model}(b) is our colour quantisation model. The motivation of our colour quantisation model architecture is driven by the IB color naming model \cite{zaslavsky2018efficient}. As illustrated in (b), the speaker represents the CQFormer, and the listener represents the classifier. We focus on the case where a colour quantiser (CQFormer) and a classifier communicate about colours. The CQFormer has a "mental" representation, \textit{i.e.} a full-colour image $\boldsymbol{x}$ associated with a prior label $y$ drawn from a prior distribution, and communicates this representation by encoding it into a colour-quantised image $\Bar{\boldsymbol{x}}$ according to the perceptual similarity. The classifier receives $\Bar{\boldsymbol{x}}$ and attempts to infer from it the full-colour image’s label $y$ by predicting the label $\hat{y}$ and constructing another distribution that approximates $y$.

There exists a problem that both the speaker and listener in Fig. \ref{fig:IB_model} (a)  have a knowledge of colour naming/recognition at the same time. In contrast, the untrained CQFormer and classifier lack knowledge of colour quantisation and image classification. Therefore, we jointly train both the CQFormer and classifier simultaneously to add prior knowledge of colour quantisation and image classification under a specific bit of colour. Finally, similar to the theoretical limit of semantic efficiency in \cite{zaslavsky2018efficient}, we obtain optimal image classification accuracy under the specific bit of colour.

\begin{figure*}[h!]
\centering
\includegraphics[width=1.0\linewidth]{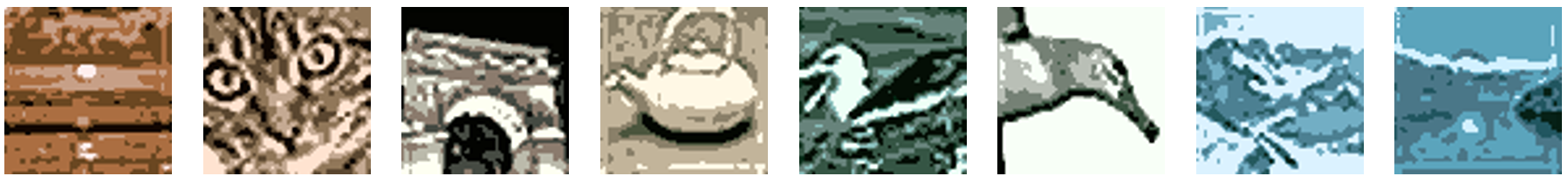}
\vspace{-5mm}
\caption{8-colour quantised images by CQFomer.}
\vspace{-6mm}
\label{fig:colourful}
\end{figure*}

\begin{figure*}[h!]
 \centering
 \includegraphics[width=1.0\linewidth]{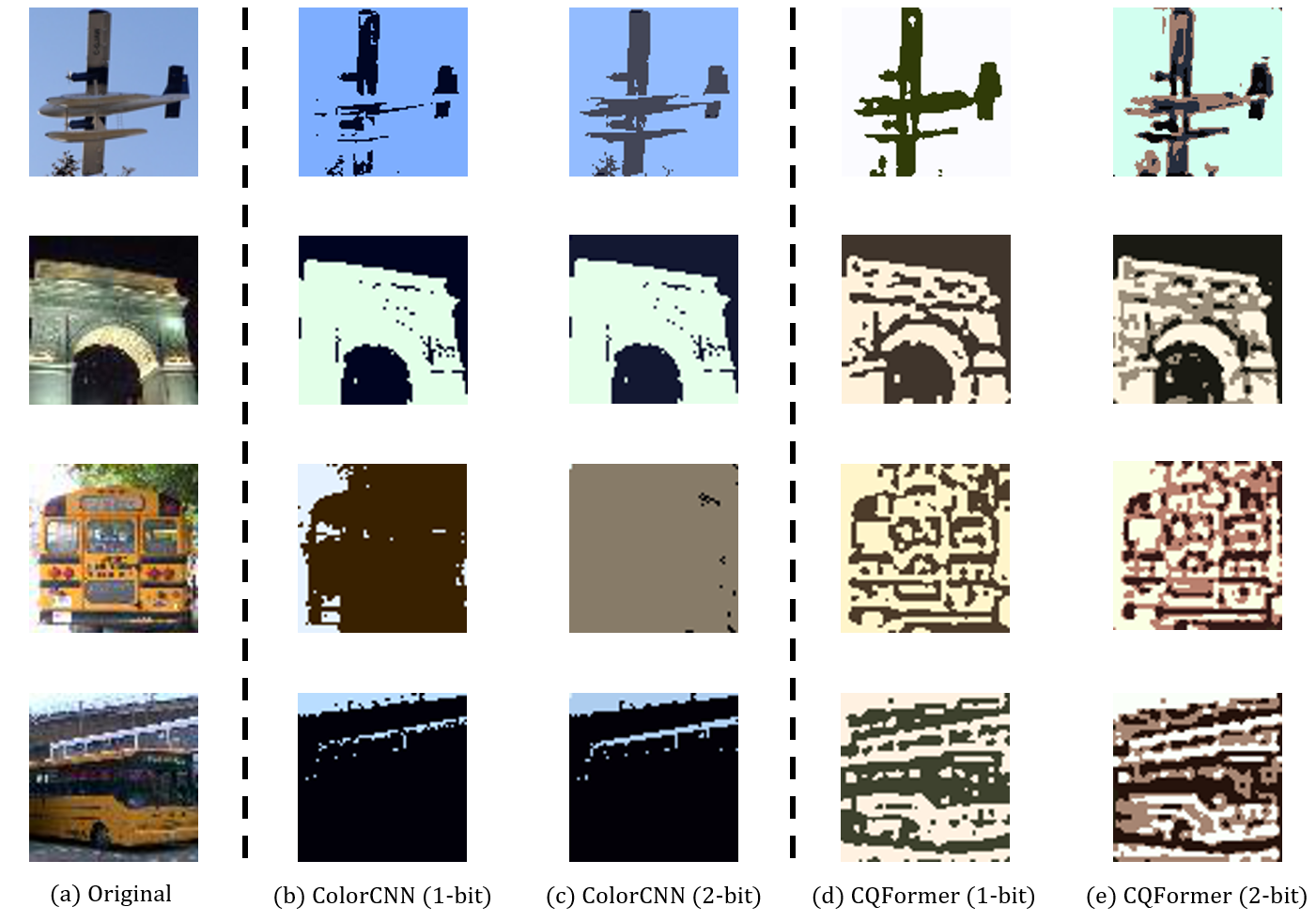}

 \caption{Visualisation of 1-bit and 2-bit colour quantisation by ColorCNN \cite{hou2020learning} and CQformer.}
 \label{fig:visualise}
\end{figure*}
\section{Visualisation}
\label{sec:Visualisation}
As shown in Fig.~\ref{fig:visualise} and Fig.~\ref{fig:colourful}, our CQFormer effectively preserves more perceptual structure and similarity. For instance, aeroplane wings, textures of architecture and vehicle windows.

{\small
\bibliographystyle{ieee_fullname}
\bibliography{egbib}
}